%% file: 00-main.tex
\pdfoutput=1

\documentclass[11pt]{article}

\usepackage[]{acl}

\usepackage{times}
\usepackage{latexsym}

\usepackage{graphicx}
\usepackage{multirow}
\usepackage{booktabs}
\graphicspath{ {./figures/} }

\usepackage[T1]{fontenc}
\usepackage[utf8]{inputenc}

\usepackage{CJKutf8}
\newcommand{\zhjp}[1]{\begin{CJK}{UTF8}{ipxm}#1\end{CJK}}

\usepackage[utf8]{inputenc}
\usepackage{xcolor,colortbl}
\definecolor{Gray}{gray}{0.85}
\newcolumntype{g}{>{\columncolor{Gray}}c}
\usepackage{soul}

\usepackage{microtype}
\usepackage{todonotes}
\newcommand*\iftodonotes{\if@todonotes@disabled\expandafter\@secondoftwo\else\expandafter\@firstoftwo\fi}
\makeatother

\title{ZusammenQA: Data Augmentation with Specialized Models for Cross-lingual Open-retrieval Question Answering System}

\author{Chia-Chien Hung\textsuperscript{1}, Tommaso Green\textsuperscript{1},  Robert Litschko\textsuperscript{1}, \\\textbf{Tornike Tsereteli\textsuperscript{1}}, \textbf{Sotaro Takeshita\textsuperscript{1}}, \textbf{Marco Bombieri\textsuperscript{2}},\\ \textbf{Goran Glava\v{s}\textsuperscript{3}} and \textbf{Simone Paolo Ponzetto\textsuperscript{1}} \\
  \textsuperscript{1} Data and Web Science Group, University of Mannheim, Germany \\
  \textsuperscript{2} ALTAIR Robotics Lab, University of Verona, Italy \\ 
  \textsuperscript{3} CAIDAS, University of Würzburg, Germany \\ 
  \texttt{\{chia-chien.hung, tommaso.green, robert.litschko,} \\
  \texttt{tornike.tsereteli, sotaro.takeshita, ponzetto\}@uni-mannheim.de}  \\ 
  \texttt{marco.bombieri\_01@univr.it}, \quad \texttt{goran.glavas@uni-wuerzburg.de}}

\begin{document}
\maketitle
\begin{abstract}
This paper introduces our proposed system for the MIA Shared Task on Cross-lingual Open-retrieval Question Answering (COQA). In this challenging scenario, given an input question the system has to gather evidence documents from a multilingual pool and generate from them an answer in the language of the question. We devised several approaches combining different model variants for three main components: \textit{Data Augmentation}, \textit{Passage Retrieval}, and \textit{Answer Generation}. 
For passage retrieval, we evaluated the monolingual BM25 ranker against the ensemble of \textit{re-rankers based on multilingual pretrained language models} (PLMs) and also variants of the shared task baseline, re-training it from scratch using a recently introduced contrastive loss that maintains a strong gradient signal throughout training by means of mixed negative samples.
For answer generation, we focused on language- and domain-specialization by means of continued language model (LM) pretraining of existing multilingual encoders.
Additionally, for both passage retrieval and answer generation, we augmented the training data provided by the task organizers with automatically generated question-answer pairs created from Wikipedia passages to mitigate the issue of data scarcity, particularly for the low-resource languages for which no training data were provided. Our results show that language- and domain-specialization as well as data augmentation help, especially for low-resource languages. 
\end{abstract}
\section{Introduction}
\input{01-intro}

\section{Data Augmentation}
\input{02-data}

\section{Methodology}
\input{03-methodology}

\section{Experimental Setup}
\label{s:setup}
\input{04-experiments}

\section{Results and Discussion}
\input{05-results}

\section{Related Work}
\input{06-rw}

\section{Reproducibility}
\input{07-reproducibility}

\section{Conclusion}
\input{08-conclusion}




\bibliography{anthology,custom}
\bibliographystyle{acl_natbib}

\appendix
\input{09-appendix}

\end{document}

%% file: 01-intro.tex
\label{sec:intro}

Open-retrieval Question Answering (OQA), where the agent helps users to retrieve answers from large-scale document collections with given \textit{open} questions, has arguably been one of the most challenging natural language processing (NLP) applications in recent years~\citep[e.g.,][]{bib:lewis, karpukhin-etal-2020-dense, izacard-grave-2021-leveraging}. As is the case with the vast majority of NLP tasks, much of the OQA focused on English, relying on a pipeline that crucially depends on a neural passage retriever, i.e., a \mbox{(re-)ranking model} -- trained on large-scale English QA datasets -- to find evidence passages in English \cite{bib:lewis} for answer generation. 
Unlike in many other retrieval-based tasks, such as ad-hoc document retrieval \cite{craswell2021ms}, parallel sentence mining \cite{zweigenbaum2018overview}, or Entity Linking \cite{wu-etal-2020-scalable},
 the progress toward \textit{Cross-lingual Open-retrieval Question Answering (COQA)} has been hindered by the lack of efficient integration and consolidation of knowledge expressed in different languages \cite{bib:1}. 
COQA is especially relevant for opinionated information, such as news, blogs, and social media. In the era of fake news and deliberate misinformation, training on only (or predominantly) English texts is more likely to lead to more biased and less reliable NLP models.
Further, an Anglo- and Indo-European-centric NLP \cite{joshi2020state} is unrepresentative of the needs of the majority of the world's population (e.g., Mandarin and Spanish have more native speakers than English, and Hindi and Arabic come close) and contributes to the widening of the digital language divide.\footnote{\url{http://labs.theguardian.com/digital-language-divide/}}   
Developing solutions for cross-lingual open QA (COQA) thus contributes towards the goal of global equity of information access.

\begin{figure*}[t]
    \centering
    \includegraphics[width=1.0\textwidth]{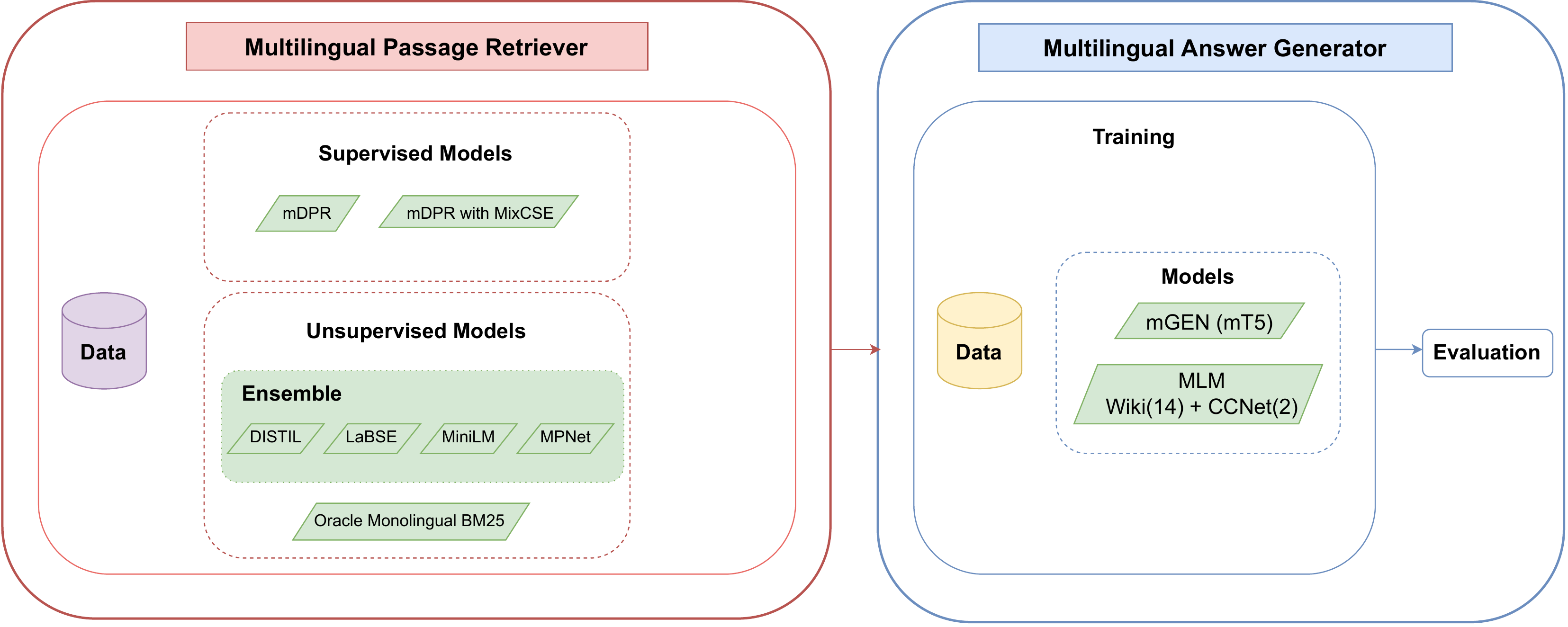}
    \caption{The proposed pipeline for the cross-lingual QA problem. The pipeline is composed of two stages: (i) the retrieval of documents containing a possible answer (red box) and the generation of an answer (blue box). For the retrieval part, we exploited different methods, based both on training mDPR variants and on the ensembling of \textit{blackbox} models. For training the mDPR variants, we enlarge the original training dataset with samples from our data augmentation pipeline. For the generation part, we enrich the existing baseline method with data augmentation and masked language modeling.}
    \label{fig:pipeline}
\end{figure*}

COQA is the task of automatic question answering, where the answer is to be found in a large multilingual document collection. 
It is a challenging NLP task, with questions written in a user's preferred language, where the system needs to find evidence in a large-scale document collection written in different languages; the answer then needs to be returned to the user in their preferred language (i.e., the language of the question).

More formally, the goal of a COQA system is to find an answer $a$ to the query $q$ in the collection of documents $\{D\}^N_i$. In the cross-lingual setting, $q$ and $\{D\}^{N}_i$ are, in general, in different languages. For example, the users can ask a question in Japanese, and the system can search an English document for an answer that then needs to be returned to the user in Japanese. 
\label{sec:method}

In this work, we propose data augmentation for specialized models that correspond to the two main components of a standard COQA system -- \textit{passage retrieval} and \textit{answer generation}: (1) we first extract the passages from all documents of all languages, exploiting both \textit{unsupervised} and \textit{supervised} (e.g., mDPR variants) passage retrieval methods; (2) we then use the retrieved passages from the previous step and further conduct intermediate training of a pretrained language model (e.g., mT5~\citep{xue-etal-2021-mt5}) on the extracted augmented data in order to inject language-specific knowledge into the model, and then generate the answers for each question from different language portions. As a result, we obtain a specialized model trained on the augmented data for the COQA system. The overall process is illustrated in Figure~\ref{fig:pipeline}.

%% file: 02-data.tex
\label{sub:data}
We use a language model to generate question-answer (QA) pairs from English texts, which we then filter according to a number of heuristics and translate into the other 15 languages.\footnote{Arabic(\textsc{ar}), Bengali(\textsc{bn}), Finish(\textsc{fi}), Japanese(\textsc{ja}), Korean(\textsc{ko}), Russian(\textsc{ru}), Telugu(\textsc{te}), Spanish(\textsc{es}), Khmer(\textsc{km}), Malay(\textsc{ms}), Swedish(\textsc{sv}), Turkish(\textsc{tr}), Chinese(\textsc{zh-cn}), Tagalog(\textsc{tl}) and Tamil(\textsc{ta}).}
An example can be seen in Figure~\ref{fig:data-aug-example} in the Appendix.

\subsection{Question-Answer Generation}\label{subsub:qa-generation}
For generating the question-answer pairs, we use the provided Wikipedia passages as the input to a language model, which then generates questions and answers based on the input text.
We based our choice of the model on the findings by \citet{NEURIPS2019_c20bb2d9} and \citet{10.5555/3524938.3524998}, who showed that language models that are fine-tuned jointly on Question Answering and Question Generation, outperform individual models fine-tuned independently on those tasks.
More specifically, we use the model by \citet{dugan2022feasibility} and make slight modifications. \citet{dugan2022feasibility} used a T5 model fine-tuned on SQuAD and further fine-tuned it on three tasks simultaneously: Question Generation (GQ), Question Answering (QA), and Answer Extraction (AE).
They also included a summarization module to create lexically diverse question-answer pairs.
We found that using this module sometimes leads to factually incorrect passages, and leave this to future work.
Similar to \citet{dugan2022feasibility}, we split the original passages into whole sentences that are shorter than 512 tokens.\footnote{We use a different sentence splitting method, namely pySBD.}
We then generate the pairs using the first three sub-passages.

\subsection{Filtering}\label{subsub:filtering}
Before translating question-answer pairs, to ensure better translations, we enforce each pair to satisfy at least one of a number of heuristics, which we determined through manual evaluation of the generated pairs.
Each pair is evaluated on whether one of the following is true (in the respective order): the answer is a number, the question starts with the word \textit{who}, the question starts with the words \textit{how many}, or the answer contains a number or a date.
After filtering, we are left with roughly 339,000 question-answer pairs.

\subsection{Translation}\label{subsub:translation}
We use the Google Translate API provided by \textit{translatepy}\footnote{\url{https://github.com/Animenosekai/translate}} to translate the filtered question-answer pairs from English into the relevant 15 languages.
Each language has an equal number of question-answer pairs.
In total, we generate about 5.4 million pairs for all languages combined.

%% file: 03-methodology.tex
Following the approach described in \citet{bib:AsaiYKH21}, we consider the COQA problem as two sub-components: the \textit{retrieval} of documents containing a possible answer and the \textit{generation} of an answer. Figure~\ref{fig:pipeline} summarizes the proposed methods.
This section is organized as follows: we present the proposed retrieval methods in \S\ref{sub:retrieval} and demonstrate the language-specialized methods for answer generation in \S\ref{sub:mgen}.

\subsection{Passage Retrieval}\label{sub:retrieval}
For the passage retrieval phase, we explored the approaches described in the following sections, which fall into three main categories: the enhancement of the training procedure of the mDPR baseline, the ensembling of \textit{blackbox} models (i.e. retrieval using multilingual sentence encoders trained for semantic similarity) and lexical retrieval using BM25.
While the first category is a \textit{supervised} approach, which uses QA datasets to inject task knowledge into pretrained models, the others use general linguistic knowledge for retrieving (i.e., \textit{unsupervised}).

\paragraph{Baseline: mDPR}
We take as a baseline the method proposed in \citet{bib:AsaiYKH21}. They propose mDPR (Multilingual Dense Passage Retriever), a model that extends the Dense Passage Retriever (DPR) \cite{dpr} to a multilingual setting. It is made of two mBERT-based encoders~\citep{devlin-etal-2019-bert}, one for the question and one for the passages. The training approach proceeds over two subsequent stages: (i) \textit{parameter updates} and (ii) \textit{cross-lingual data mining}.

In the first phase, both mDPR and mGEN (\S\ref{para:mgen}) are trained one after the other. For mDPR, the model processes a dataset  $\mathcal{D} = \{ (q_i, p_i^+, p_{i,1}^{-}, p_{i,2}^{-}, \dots, p_{i,n}^{-} ) \}_{i=1}^m $ made of tuples containing a question $q_i$, the passage $p_i^+$ containing an answer (called positive or gold), and a set $\{ p_{i,j}^-\}_{j=1}^{n}$ of negative passages. For every question, negatives are made up of the positive passages from the other questions or passages either extracted at random or produced by the subsequent data mining phase.
To do this, they use a contrastive loss ($\mathcal{L}_{\text{mdpr}}$) that moves the embedding of the question close to its positive passage, while at the same time repelling the representations of negative passages:

\begin{equation*} 
  \mathcal{L}_{\text{mdpr}} = - \log  \frac{\langle \mathbf{e}_{q_i},  \mathbf{e}_{p_i^+} \rangle}{\langle \mathbf{e}_{q_i},  \mathbf{e}_{p_i^+} \rangle + \sum_{j=1}^{n} \langle \mathbf{e}_{q_i},  \mathbf{e}_{p_{i,j}^-} \rangle} 
\end{equation*}

In the second stage, the training set is expanded by finding new positive and negative passages using  Wikipedia language links and mGEN (\S\ref{para:mgen}) to automatically label passages. This two-staged training pipeline is repeated $T$ times.

\paragraph{mDPR variants}
\label{sub:mdpr_mxicse}
One of our approaches is to simply substitute the loss function presented above with a contrastive loss described in \citet{mixloss}, named MixCSE. 
 In this work, the authors tackle a common problem of out-of-the-box BERT sentence embeddings, called anisotropy~\citep{li-etal-2020-sentence}, which makes all the sentence representations to be distributed in a narrow cone. Contrastive learning has already proven effective in alleviating this issue by distributing embeddings in a larger space \cite{gao-etal-2021-simcse}. 
 \citet{mixloss} prove that hard negatives, i.e. data points hard to distinguish from the selected anchor, are key for keeping a strong gradient signal; however, as learning proceeds, they become orthogonal to the anchor and make the gradient signal close to zero.
 For this reason, the key idea of MixCSE is to continually generate hard negatives via mixing positive and negative examples, which maintains a strong gradient signal throughout training. Adapting this concept to our retrieval scenario, we construct mixed negative passages as follows:
 \[
   \tilde{\mathbf{e}_i} = \frac{\lambda \, \mathbf{e}_{p_i^+} + (1 - \lambda) \, \mathbf{e}_{p_{i,j}^-}}{\lVert \lambda \, \mathbf{e}_{p_i^+} + (1 - \lambda) \, \mathbf{e}_{p_{i,j}^-} \rVert_2},
 \]
 
 where $p_{i,j}^-$ is a negative passage chosen at random. 
 We provide the equation of the loss in Appendix~\ref{app:mixcse}. The main difference with $\mathcal{L}_{\text{mdpr}}$ is the addition of a mixed negative in the denominator and the similarity used (exponential of cosine similarity instead of dot product). 
 
We train mDPR with the original loss and with the MixCSE loss on the concatenation of the provided training set for mDPR and the augmented data obtained via the methods described in \S\ref{sub:data}. We refer to these two variants as \textit{mDPR(AUG)} and \textit{mDPR(AUG) with MixCSE}, respectively.

\paragraph{Ensembling ``blackbox'' models}
Following the approaches presented in \citet{Litschko2021EvaluatingMT}, we also ensemble the ranking of some \textit{blackbox} models that directly produce a semantic embedding of the input text. We provide a brief overview of the models included in our ensemble below. 
\begin{itemize}
    \item \textbf{DISTIL} \cite{distil} is a teacher-student framework for injecting the knowledge obtained through specialization for semantic similarity from a specialized monolingual transformer (e.g., BERT) into a non-specialized multilingual transformer (e.g., mBERT).
For semantic similarity, it first specializes a monolingual (English) teacher encoder using the available semantic sentence-matching datasets for supervision. In the second knowledge distillation step, a pretrained multilingual student encoder is trained to mimic the output of the teacher model.
We benchmark different DISTIL models: 
\begin{itemize}
    \item DISTIL\textsubscript{use}: instantiates the student as the pretrained m-USE~\citep{yang-etal-2020-multilingual} instance;
    \item DISTIL\textsubscript{xlmr}: initializes the student model with the pretrained XLM-R \cite{conneau-etal-2020-unsupervised} transformer;
    \item DISTIL\textsubscript{dmbert}: distills the knowledge from the Sentence-BERT \cite{reimers-gurevych-2019-sentence} teacher into a multilingual version of DistilBERT~\citep{DBLP:journals/corr/abs-1910-01108}, a 6-layer transformer pre-distilled from mBERT.
\end{itemize}

\item \textbf{LaBSE} (Language-agnostic BERT Sentence Embeddings \citet{labse}) is a neural dual-encoder framework, trained with parallel data. LaBSE training starts from a pretrained mBERT instance. LaBSE additionally uses standard self-supervised objectives used in the pretraining of mBERT and XLM~\citep{conneau2019cross}: masked and translation language modelling (MLM and TLM). 

\item \textbf{MiniLM} \cite{miniml} is a student model trained by deeply mimicking the self-attention behavior of the last Transformer layer of the teacher, which allows a flexible number of layers for the students and alleviates the effort of finding the best layer mapping.

\item \textbf{MPNet} \cite{mpnet} is based on a pre-training method that leverages the dependency among the predicted tokens through permuted language modeling and makes the model see auxiliary position information to reduce the discrepancy between pre-training and fine-tuning.
\end{itemize}

We produce an ensembling of the \textit{blackbox} models by simply taking an average of the ranks for each of the documents retrieved, which is denoted as \textit{EnsembleRank}.

\paragraph{Oracle Monolingual BM25} \cite{bm25_1, bm25_2}
This approach is made of two phases: first, we automatically detect the language of the question, then we query the index in the detected language.
As a weighting scheme in the vector space model, we choose BM25. It is based on a probabilistic interpretation of how terms contribute to the document's relevance. It uses exact term matching and the score is derived from a sum of contributions from each query term that appears in the document.
We use an \emph{oracle BM25} approach: this naming derives from the fact that we query the index with the answer rather than the question. This was done at training time to increase the probability of the answer to be in the passages consumed by mGEN, so that the generation model would hopefully learn to extract the answer from its input, rather than generating it from the question only. At inference time, we query the index using the question.

\begin{table*}[t]
\resizebox{\textwidth}{!}{%
\begin{tabular}{l|l|l}
\toprule
 &
  \textbf{AUG-QA} &
  \textbf{AUG-QAP} \\ \midrule
\textbf{QA pair} &  \multicolumn{2}{|l}{\begin{tabular}[c]{@{}l@{}} 
 \zhjp{\textbf{Q}: レゴグループを設立したのは誰ですか？}\\\zhjp{\textbf{A}: オレ・カーク・クリスチャンセン}\end{tabular}}
 \\
 \midrule
  \textbf{Passage} &
  \begin{tabular}[c]{@{}l@{}}{}The Lego Group began manufacturing the interlocking toy bricks in 1949. \\  Movies, games, competitions, and six Legoland amusement parks have been \\
  developed under the brand. As of July 2015, 600 billion Lego parts had been \\ produced. In February 2015, Lego replaced Ferrari as Brand Finance's \\ ``world's most powerful brand''. History. The Lego Group began in the \\ workshop of Ole Kirk Christiansen  \end{tabular} &
  \begin{tabular}[c]{@{}l@{}}\zhjp{レゴグループは1949年に連動おもちゃのレンガの製造を開始しました。映画、} \\
 \zhjp{ゲーム、競技会、および6つのレゴランド遊園地がこのブランドで開発されま} \\   \zhjp{した。2015年7月現在、6,000 億個のレゴパーツが生産されています。2015年}\\
 \zhjp{2月、レゴはブランドファイナンスの「世界で最も強力なブランド」としてフ}\\  
 \zhjp{ェラーリに取って代わりました。歴史。レゴグループは、OleKirkChristiansen}\\
 \zhjp{のワークショップで始まりました
} \\ \end{tabular} \\ 
 
 \bottomrule
\end{tabular}
}
\caption{Examples of augmented training instances for \textbf{AUG-QA} and \textbf{AUG-QAP}. Top row: translated question-answer pair in Japanese. Below are different training examples: (1) \textbf{AUG-QA}: the English Wikipedia passage is kept with the translated question-answer pair. (2) \textbf{AUG-QAP}: the English Wikipedia passage is translated to the same language as the question-answer pair.}
\label{tab:QAP}
\end{table*}

\subsection{Answer Generation}\label{sub:mgen}
Our answer generation modules take a concatenation of the question and the related documents retrieved by the retrieval module as an input and generate an answer.
In this section, we first explain the baseline system which is the basis of our proposed approaches and then present our specialization method.

\paragraph{Baseline: mGEN}
\label{para:mgen}
We use mGEN (Multilingual Answer Generator; \citet{bib:AsaiYKH21}) as the baseline for the answer generation phase.
They propose to take mT5 \citep{xue-etal-2021-mt5}, a multilingual version of a pretrained transformer-based encoder-decoder model \citep{raffel-etal-2020-t5}, and fine-tune it for multilingual answer generation.
The pre-training process of mT5 is based on a variant of masked language modeling named span-corruption, in which the objective is to reconstruct continuously masked tokens in an input sentence~\citep{xue-etal-2021-mt5}.
For fine-tuning, the model is trained on a sequence-to-sequence (seq2seq) task as follows:
\begin{equation*}
  P(a^L \, | \, q^L, P^{N}) = \prod_{i}^{T} p(a_{i}^{L}|a_{<i}^{L}, q^{L}, P^{N})
\end{equation*}
The model predicts a probability distribution over its vocabulary at each time step ($i$). It is conditioned on the previously generated answer tokens ($a_{<i}^{L}$), the input question ($q^L$) and $N$ retrieved passages ($P^N$).
Because of a possible language mismatch between the answer and the passages, it is not possible to extract answers as in existing work in monolingual QA tasks~\citep{karpukhin-etal-2020-dense}: for this reason, mGEN opts for directly generating answers instead.

\paragraph{Masked Language Modeling (MLM)}
Following successful work on language-specialized pre-training via language modeling~\citep{glavas-etal-2020-xhate, hung-etal-2022-multi2woz}, we investigate the effect of running MLM on the language-specific portions of Wikipedia passages~\citep{bib:AsaiYKH21} and {CCN}et~\citep{wenzek-etal-2020-ccnet} with mT5~\citep{xue-etal-2021-mt5}. For the extracted texts of all 16 languages, 14 languages are from the released Wikipedia passages and the missing two \textit{surprise} languages (Tamil, Tagalog) are from {CCN}et. We additionally clean all language portions by removing email addresses, URLs, extra emojis and punctuations, and selected 7K for training and 0.7K for validation for each language. In this way, we inject both the domain-specific (i.e., Wikipedia knowledge) and language-specific (i.e., 16 languages) knowledge into the multilingual pretrained language model via MLMing as an intermediate specialization step.

\paragraph{Augmentation Data Variants}
To further investigate model capability on (1) extracting answers from English passages or (2) extracting answers from translated passages, while keeping the Question-Answer pairs in other non-English languages, we conduct experiments on two augmentation data variants: \textbf{AUG-QA} and \textbf{AUG-QAP}. \textbf{AUG-QA} keeps the English passage with the translated Question-Answer pairs, while \textbf{AUG-QAP} translates the English passage to the same language as the translated Question-Answer pairs. Detailed examples are shown in Table~\ref{tab:QAP}.

%% file: 04-experiments.tex
\label{sec:experiments}
We demonstrate the effectiveness of our proposed COQA systems by comparing them to the baseline models and thoroughly comparing different specialization methods from \S\ref{sec:method}.

\setlength{\tabcolsep}{20pt}
\begin{table}[t]
\centering
\scriptsize{
\begin{tabular}{l|c|c} 
\toprule
\textbf{Dataset} & \multicolumn{1}{l|}{\textbf{Lang}} & \multicolumn{1}{l}{\textbf{Train size}} \\ 
\midrule
Natural Questions & en & 76635 \\ 
\midrule
\multirow{7}{*}{XOR-TyDi QA} & ar & 18402 \\
 & bn & 5007 \\
 & fi & 9768 \\
 & ja & 7815 \\
 & ko & 4319 \\
 & ru & 9290 \\
 & te & 6759 \\
\bottomrule
\end{tabular}
}
\caption{The training data size for 8 languages.}
\label{tab:train_data}
\end{table}

\setlength{\tabcolsep}{13pt}
\begin{table}[t!]
\centering
\scriptsize{
\begin{tabular}{l|c|c|c}
\toprule
\textbf{Dataset} & \textbf{Lang} & \textbf{Dev size}& \textbf{Test size} \\ 
\midrule 
MKQA (parallel) & 12 & 1758 & 5000 \\ 
\midrule
\multirow{7}{*}{XOR-TyDi QA} & ar & 590 & 1387 \\
 & bn & 203 & 490 \\
 & fi & 1368 & 974 \\
 & ja & 1056 & 693 \\
 & ko & 1048 & 473 \\
 & ru & 910 & 1018 \\
 & te & 873 & 564 \\ 
\midrule
\multirow{2}{*}{Surprise} & ta & - & 350 \\
 & tl & - & 350 \\
\bottomrule
\end{tabular}
}
\caption{The development and test data size for each language. The  data size for MKQA is equal for all 12 languages. Two \textit{surprise} languages are provided without development data.}
\label{tab:dev_test_data}
\end{table}

\paragraph{Evaluation Task and Measures}
Our proposed approaches are evaluated in 16 languages, 8 of which are not covered in the training data.\footnote{Languages with training data: English(\textsc{en}), Arabic(\textsc{ar}), Bengali(\textsc{bn}), Finish(\textsc{fi}), Japanese(\textsc{ja}), Korean(\textsc{ko}), Russian(\textsc{ru}), Telugu(\textsc{te}). Without training data: Spanish(\textsc{es}), Khmer(\textsc{km}), Malay(\textsc{ms}), Swedish(\textsc{sv}), Turkish(\textsc{tr}), Chinese(\textsc{zh-cn}). Tagalog(\textsc{tl}) and Tamil(\textsc{ta}) are considered as \textit{surprise} languages.} The training and evaluation data are originally from Natural Questions~\cite{kwiatkowski-etal-2019-natural}, XOR-TyDi QA~\cite{bib:datasetAsaiKCLCH21}, and MKQA~\cite{bib:MKQA}. Data size statistics for each resource and language are shown in Table~\ref{tab:train_data} and \ref{tab:dev_test_data}.

The evaluation results are measured on the competition platform hosted at eval.ai.\footnote{\url{https://eval.ai/}} The systems are evaluated on two COQA datasets: XOR-TyDi QA \cite{bib:datasetAsaiKCLCH21}, and MKQA \cite{bib:MKQA}, using token-level F1 (\textbf{F1}), as common evaluation practice of open QA systems~\citep{lee-etal-2019-latent}. For \textit{non-spacing} languages, we follow the token-level tokenizers~\footnote{Tokenizers for non-spacing languages: Mecab (\textsc{ja}); khmernltk (\textsc{km}); jieba (\textsc{zh-cn}).} for both predictions and ground-truth answers. The overall score is calculated by using macro-average scores on XOR-TyDi QA and MKQA datasets, and then taking the average F1 scores of both datasets.


\paragraph{Data}
\label{sec:data}
We explicitly state that we did not train on the development data or the subsets of the Natural Questions and TyDi QA, which are used to create MKQA or XOR-TyDi QA datasets. This makes all of our proposed approaches fall into the \textit{constrained} setup proposed by the organizers.

For training the mDPR variants, we exploit the organizer's dataset that was obtained from DPR Natural Questions \cite{dpr} and XOR-TyDiQA gold paragraph data. More specifically, for training and validation, we always use the version of the dataset containing augmented positive and negative passages obtained from the top 50 retrieval results of the organizer's mDPR.
We merge this dataset with the augmented data, filtering the latter to get $100$k samples for each of the $16$ languages.

We base our training data for answer generation models on the organizer's datasets with the top 15 retrieved documents from the coupled retriever.
To use automatically generated question-answer pairs for each language from \S\ref{sec:data} for fine-tuning, we align the format with retrieved results by randomly sampling passages from English Wikipedia as negative contexts,\footnote{For the negative contexts, we use the passages that were used for generating the question-answer pairs (i.e., the first three sub-passages). These were then trimmed down to 100 tokens. We ensure that the answer is not contained in the negative contexts through lowercase string-matching.} while we keep the seed documents as positive ones.
We explore two ways of merging the positive and negative passages: in the "shuffle" style, the positive passage appears in one of the top 3 documents; in the "non-shuffle" method, the positive passage always appears on the top. However, since these two configurations did not show large differences, we only report the former one in this paper.
We also investigated if translating passages into the different 16 languages
\footnote{Using the same Google Translate API adopted for the QA translation in \S\ref{sec:data}.} may be beneficial with respect to keeping all the passages in English (\textbf{AUG-QA}).
Due to computational limitations, in our data augmented setting for generation model fine-tuning, we use 2K question-answer pairs with positive/negative passages for each language for our final results.

\paragraph{Hyperparameters and Optimization}

For multilingual dense passage retrieval, we mostly follow the setup provided by the organizers: learning rate $1e-5$ with AdamW \citep{loshchilov2018decoupled}, linear
scheduling with warm-up for $300$ steps and dropout rate $0.1$. For \textbf{mDPR(AUG) \textit{with} MixCSE} \cite{mixloss}, we use $\lambda = 0.2$ and $\tau=0.05$ for the loss (see Appendix~\ref{app:mixcse}).
We train with a batch size of $16$ on $1$ GPU for at most $40$ epochs, using average rank on the validation data to pick the checkpoints. The training is done independently of mGEN, in a non-iterative fashion.

For retrieving the passages, we use cosine similarity between question and passage across all proposed retrieval models, returning the top $100$ passages for each of the questions.

For language-specialized pretraining via MLM, we use AdaFactor \citep{shazeer2018adafactor} with the learning rate $1e-5$ and linear scheduling with warm-up for $2000$ steps up to 20 epochs.
For multilingual answer generation fine-tuning, we also mostly keep the setup from the organizers: learning rate $3e-5$ with AdamW \citep{loshchilov2018decoupled}, linear scheduling with warm-up for $500$ steps, and dropout rate as $0.1$. We take the top 15 documents from the retrieved results as our input and truncate the input sequence after $16,000$ tokens to fit the model into the memory constraints of our available infrastructure.


%% file: 05-results.tex
\label{sec:results}

\setlength{\tabcolsep}{11pt}
\begin{table*}[t]
\centering
\scriptsize{
\begin{tabular}{l cccccccg}
\toprule
\vspace{0.3em}
 & \multicolumn{8}{c}{\textbf{XOR-TyDi QA}} \\
\textbf{Models} & \textbf{ar} & \textbf{bn} & \textbf{fi} & \textbf{ja} & \textbf{ko} & \textbf{ru} & \textbf{te} & \textbf{Avg.} \\\cmidrule(lr){2-8}
mDPR + mGEN (baseline 1) & 49.66 & 33.99 & \textbf{39.54} & 39.72 & \textbf{25.59} & \textbf{40.98} & 36.16 & 37.949\\
\midrule
\textit{Unsupervised Retrieval} &  &  &  &  &  &  &  & \multicolumn{1}{c}{}\\
OracleBM25 + MLM-14   & 0.34 & 0.49 & 0.52 & 2.56  & 0.19 & 0.57 & 5.16 & 1.404 \\
EnsembleRank + MLM-14  &  0.34& 0.49 & 1.33 & 2.56 & 0.38 & 6.27 & 16.21 & 3.161 \\
\midrule
\textit{Supervised Retrieval} &  &  &  &  &  &  &  & \multicolumn{1}{c}{}\\
mDPR(AUG) \textit{with} MixCSE + MLM-14 & 20.94 & 7.18 & 15.27 & 23.16 & 10.25 & 19.23 & 10.53 & 15.223 \\
mDPR(AUG) + MLM-14 & 24.99 & 15.19 & 20.33 & 22.31 & 10.68 &18.82  &11.97  &17.754  \\
mDPR + MLM-14 & \textbf{51.66} & 31.96 & 38.68 & 40.89 & 25.35 & 39.87 & \textbf{37.26} & \textbf{37.951}\\
mDPR + MLM-14(XORQA \& AUG-QA)  &  49.41 & 32.90 & 37.95 & \textbf{40.97} & 24.22 & 39.29 & 35.76 & 37.213\\
mDPR + MLM-14(XORQA \& AUG-QAP)  &  48.79&  33.73& 38.33 & 39.87 & 25.26 & 39.11 & 37.94 & 37.577 \\
mDPR + MLM-16  & 49.92 & 31.16 & 37.20 & 39.92 & 24.63 & 38.78 & 34.30 & 36.558 \\
mDPR + MLM-16(XORQA \& AUG-QA)  & 49.45 & 31.59 & 38.33 & 40.44  & 23.83 & 38.67 & 35.92 & 36.889 \\
mDPR + MLM-16(XORQA \& AUG-QAP)  & 48.21 & \textbf{34.20} & 38.78  & 40.76 & 24.81 & 39.49 & 34.37 & 37.231\\
  \bottomrule
\end{tabular}%
}
\caption{Evaluation results on XOR-TyDi QA test data with F1 and macro-average F1 scores.}
\label{tab:xor-result}
\end{table*}

\setlength{\tabcolsep}{3.2pt}
\begin{table*}[t!]
\centering
\scriptsize{
\begin{tabular}{l cccccccccccc cc g}
\toprule
 \vspace{0.3em}
 & \multicolumn{12}{c}{\textbf{MKQA}} & \multicolumn{2}{c}{\textbf{Surprise}} & \multicolumn{1}{c}{}\\
\textbf{Models} &\textbf{ar} & \textbf{en} & \textbf{es} & \textbf{fi} & \textbf{ja} & \textbf{km} & \textbf{ko} & \textbf{ms} & \textbf{ru} & \textbf{sv} & \textbf{tr} & \textbf{zh-cn} & \textbf{ta} & \textbf{tl} &  \textbf{Avg.} \\ \cmidrule(lr){2-13}\cmidrule(lr){14-15}
mDPR + mGEN (baseline1)&  \textbf{9.52} & \textbf{36.34} & \textbf{27.23} & \textbf{22.70} & \textbf{15.89} & 6.00 & \textbf{7.68} &  \textbf{25.11} &  \textbf{14.60} & \textbf{26.69} & \textbf{21.66} & 13.78 & 0.00& 12.78& \textbf{17.141} \\
\midrule
\textit{Unsupervised Retrieval} &  &  &  &  &  &  &  &  &  &  &  &  &  & &\multicolumn{1}{c}{}\\
OracleBM25 + MLM-14  & 2.80 & 10.81& 3.70 & 3.29 & 5.89 & 1.53 & 1.51 & 5.49 & 1.85 & 7.42 & 2.94& 1.81 & 0.00 & 8.23 & 4.090\\
EnsembleRank + MLM-14  & 6.43 & 31.66 & 20.02 & 17.38 & 10.68 & \textbf{6.24} & 4.38 & 21.03 & 6.27 & 21.09 & 17.13&  7.22& 0.00 & 8.39 & 12.709\\
\midrule
\textit{Supervised Retrieval} &  &  &  &  &  &  &  &  &  &  &  &  &  & &\multicolumn{1}{c}{}\\
mDPR(AUG) \textit{with} MixCSE + MLM-14 & 4.71 & 28.06 & 12.78 & 8.22 &7.92  &5.44  & 2.74 & 12.90 & 4.65 & 13.86 & 8.38 & 3.99 & 0.00 & 6.72&8.599\\
mDPR(AUG) + MLM-14 & 5.64 & 29.23 & 17.27 & 15.51 &7.81  &5.83  & 3.38 & 16.57 & 6.80 & 17.21 & 13.10& 4.53 & 0.00 & 8.09&10.785\\
mDPR + MLM-14 & 8.73 & 35.32 & 25.54 & 20.42 & 14.27 & 6.06 & 6.78 & 24.10 & 12.01 & 25.97& 20.27 &  13.95 & 0.00& 11.14 & 16.040\\
 mDPR + MLM-14(XORQA \& AUG-QA) & 8.46 & 35.12 & 24.74 & 19.50 & 14.38 & 5.62 & 7.22 & 23.24 & 11.46 & 24.49 & 19.67 & \textbf{15.79} & \textbf{0.86} & 12.18 & 15.909\\
 mDPR + MLM-14(XORQA \& AUG-QAP) & 8.48 & 34.73 & 25.46 & 20.09 & 14.61 & 5.00 & 7.42 & 24.16 & 12.04 &  25.61&  19.62& 15.60 & 0.00 & 12.41& 16.089\\
mDPR + MLM-16 & 8.15 & 34.14 & 24.85 & 19.38 & 13.73 & 5.93 & 6.51 & 22.21 & 11.46 & 24.91 & 18.82& 13.62 & 0.00& 12.59&15.451 \\
 mDPR + MLM-16(XORQA \& AUG-QA) & 8.21 & 34.06 & 25.65 & 20.14 & 14.22 & 5.80 & 6.70 & 24.40 & 11.82 & 25.71 & 19.92 & 15.42 & 0.40 & 12.36 & 16.057\\
 mDPR + MLM-16(XORQA \& AUG-QAP) & 8.08 & 33.89 & 24.94 & 20.50 & 14.11 & 5.15 & 7.15 & 22.95 & 12.95 & 24.93 & 19.68 & 15.27 & 0.14 & \textbf{13.07}& 15.915\\
  \bottomrule
\end{tabular}%
}
\caption{Evaluation results on MKQA test dataset and two \textit{surprise} languages with F1 and macro-average F1 scores.}
\label{tab:mkqa-result}
\end{table*}

\setlength{\tabcolsep}{20pt}
\begin{table}[t]
\centering
\scriptsize{
\begin{tabular}{l c}
\toprule
\textbf{Models} & \textbf{Avg.} \\
\midrule
mDPR + mGEN (baseline1) & \textbf{27.55} \\
\midrule
\textit{Unsupervised Retrieval} &  \\
OracleBM25 + MLM-14 & 2.75\\
EnsembleRank + MLM-wiki14  & 7.94\\
\midrule
\textit{Supervised Retrieval} &  \\
mDPR(AUG) \textit{with} MixCSE + MLM-14 &  11.91\\
mDPR(AUG) + MLM-14 &  14.27\\
mDPR + MLM-14 & 27.00\\
mDPR + MLM-14(XORQA \& AUG-QA) &  26.56\\
mDPR + MLM-14(XORQA \& AUG-QAP) &  26.83 \\
mDPR + MLM-16 & 26.00\\
mDPR + MLM-16(XORQA \& AUG-QA) &  26.47\\
mDPR + MLM-16(XORQA \& AUG-QAP) & 26.57 \\
  \bottomrule
\end{tabular}%
}
\caption{Results of macro-average F1 for two QA datasets: XOR-TyDi QA, MKQA, and two \textit{surprise} languages.}
\label{tab:overall-result}
\end{table}

\paragraph{Results Overview}
Results in Table~\ref{tab:xor-result} show the comparison between the baseline and our proposed methods on XOR-TyDi QA
while Table~\ref{tab:mkqa-result} shows the results on MKQA.
While we can see that the additional pretraining on the answer generation model (\textit{mDPR+MLM-14}) helps to outperform the baseline in XOR-TyDi QA, the same approach leads to a degradation in MKQA. 
None of the proposed methods for the retrieval module improved over the baseline mDPR in both datasets, as shown in Table~\ref{tab:overall-result}. 

\paragraph{Unsupervised vs Supervised Retrieval}
In all evaluation settings, unsupervised retrieval methods underperform supervised methods by a large margin (see Table~\ref{tab:xor-result} and \ref{tab:mkqa-result}). This might be due to the nature of the task, which is to find a document containing an answer, rather than simply finding a document similar to the input question. For this reason, such an objective might not align well with models specialized in semantic similarity ~\citep{Litschko2021EvaluatingMT}. Fine-tuning mBERT, however, makes the model learn to focus on retrieving an answer-containing document and not simply retrieving documents similar to the question.

\paragraph{Language Specialization}
We compare the evaluation results for the fine-tuned answer generation model with and without language specialization (i.e., MLMing): for \textit{XORQA-ar} and \textit{XORQA-te} we have +2.0 and +1.1 percentage points improvement compared to the baseline model (with mT5 trained on 100+ languages). We further distinguish MLM-14 and MLM-16, where the former is trained on the released Wikipedia passages for 14 languages and the latter is trained on the concatenation of Wikipedia passages and CCNet~\citep{wenzek-etal-2020-ccnet}, to which we resort for the two \textit{surprise} languages (Tamil and Tagalog), which were missing in the Wikipedia data. Overall, MLM-14 performs better than MLM-16: we hypothesize that this might be due to the domain difference between text coming from Wikipedia and CCNet: the latter is not strictly aligned with the structured text (i.e., clean) version of Wikipedia passages, and causes a slight drop in performance as we train for $2$ additional languages.

\paragraph{Data Augmentation}
Data augmentation is considered a way to mitigate the performance of low-resource languages while reaching performance on par with high-resource languages \cite{kumar-etal-2019-cross,riabi-etal-2021-synthetic,shakeri-etal-2021-towards}. Two variations are considered: AUG-QA and AUG-QAP, while the former concatenates the XOR-Tydi QA training set with the additional augmented data with translated Question-Answer pairs, and the latter is made from the concatenation of both XOR-Tydi QA training set and the translated Question-Answer-Passage.\footnote{XORQA-Tydi QA training set is with 8 languages (see Table~\ref{tab:train_data}) and augmented data are with \emph{all} 16 languages included in the test set.} We assume that by also translating passages, the setting should be closer to test time when the retrieval module can retrieve passages in any of 14 languages (without the two surprise languages). In contrast, in AUG-QA setting, the input passages to the answer generation are always in English.  
Models trained with additional AUG-QA data could increase the capacity of \textit{seeing} more data for unseen languages, while AUG-QAP may further enhance the ability of the model to generate answers from the translated passages. 
As expected, models trained with additional augmented data have better performance compared to the ones without. The encouraging finding states that, especially for two \textit{surprise} languages, the language specialized models fine-tuned with both XOR-Tydi QA and AUG-QAP drastically improve the performance of these \textit{unseen}, \textit{low-resource} languages.


\paragraph{mDPR variants results}

As shown in Table~\ref{tab:xor-result} and~\ref{tab:mkqa-result}, we can see that the mDPR variants we trained are considerably worse than the baseline. We think this is mainly caused by the limited batch size used ($16$) which is a  constraint due to our infrastructure. The number of samples in a batch is critical for contrastive training, as larger batches provide a stronger signal due to a higher number of negatives. For this reason, we think that the mDPR variants have not been thoroughly investigated and might still prove beneficial when trained with larger batches.

%% file: 06-rw.tex


\paragraph{Passage Retrieval and Answer Generation}
To improve the information accessibility, open-retrieval question answering systems are attracting much attention in NLP applications~\citep{chen-etal-2017-reading,karpukhin-etal-2020-dense}. \citet{rajpurkar-etal-2016-squad} were one of the early works to present a benchmark that requires systems to understand a passage to produce an answer to a given question. \citep{kwiatkowski-etal-2019-natural} presented a more challenging and realistic dataset with questions collected from a search engine.
To tackle these complex and knowledge-demanding QA tasks, \citet{bib:lewis} proposed to first retrieve related documents from a given question and use them as additional aids to predict an answer. In particular, they explored a general-purpose fine-tuning recipe for retrieval-augmented generation models, which combine pretrained parametric and non-parametric memory for language generation. 
\citet{izacard-grave-2021-leveraging}, they solved the problem in two steps, first retrieving support passages before processing them with a seq2seq model, and \citet{Weiwei_sigir21} further extended to the cross-lingual conversational domain.
Some works are explored with \textit{translate-then-answer} approach, in which texts are translated into English, making the task monolingual \citep{ture-boschee-2016-learning, bib:datasetAsaiKCLCH21}.
While this approach is conceptually simple, it is known to cause the \textit{error propagation} problem in which errors of the translation get amplified in the answer generation stage \citep{zhu-etal-2019-ncls}.
To mitigate this problem, \citet{bib:AsaiYKH21} proposed to extend \citet{bib:lewis} by using multilingual models for both the passage retrieval \citep{devlin-etal-2019-bert} and answer generation \citep{xue-etal-2021-mt5}.


\paragraph{Data Augmentation}
Data augmentation is a common approach to reduce the data sparsity for deep learning models in NLP \citep{feng-etal-2021-survey}.
For Question Answering (QA), data augmentation has been used to generate paraphrases via \textit{back-translation} \cite{longpre-etal-2019-exploration}, to replace parts of the input text with translations \cite{xlda_singh_19}, and to generate novel questions or answers \cite{riabi-etal-2021-synthetic,shakeri-etal-2021-towards,dugan2022feasibility}.
In the cross-lingual setting, available data have been translated into different languages \cite{xlda_singh_19,kumar-etal-2019-cross,riabi-etal-2021-synthetic,shakeri-etal-2021-towards} and language models have been used to train question and answer generation models \cite{kumar-etal-2019-cross, Chi_Dong_Wei_Wang_Mao_Huang_2020,riabi-etal-2021-synthetic,shakeri-etal-2021-towards}.

Our approach is different from previous work in Cross-lingual Question Answering task in that it only requires English passages to augment the training data, as answers are generated automatically from the trained model by \citet{dugan2022feasibility}.
In addition, our filtering heuristics remove incorrectly generated question-answer pairs, which allows us to keep only question-answer pairs with answers that are more likely to be translated correctly, thus limiting the problem of error propagation.


%% file: 07-reproducibility.tex
To ensure full reproducibility of our results and further fuel research on COQA systems, we release the model within the Huggingface repository as the publicly available multilingual pretrained language model specialized in $14$ and $16$ languages.\footnote{MLM-14: \url{https://huggingface.co/umanlp/mt5-mlm-wiki14}; MLM-16: \url{https://huggingface.co/umanlp/mt5-mlm-16}} We also release our code and data, which make our approach completely transparent and fully reproducible.
All resources developed as part of this work are publicly available at: \url{https://github.com/umanlp/ZusammenQA}. 

%% file: 08-conclusion.tex
We introduced a framework for a cross-lingual open-retrieval question-answering system, using data augmentation with specialized models in a \textit{constrained} setup. Given a question, we first retrieved top relevant documents and further generated the answer with the specialized models (i.e., MLM-ing on Wikipedia passages) along with the augmented data variants. We demonstrated the effectiveness of data augmentation techniques with language- and domain-specialized additional training, especially for resource-lean languages. However, there are still remaining challenges, especially in the retrieval model training with limited computational resources. Our future efforts will be to focus on more efficient approaches of both multilingual passage retrieval and multilingual answer generation~\citep{abdaoui-etal-2020-load} with the investigation of different data augmentation techniques~\citep{zhu-etal-2019-ncls}. We hope that our generated QA language resources with the released models can catalyze the research focus on resource-lean languages for COQA systems.


%% file: 09-appendix.tex
\clearpage


\newpage
\onecolumn
\section{Data Augmentation Example}

Figure~\ref{fig:data-aug-example} shows an example of a Wikipedia passage about \textit{An American in Paris}.
The passage in orange is the set of sentences whose length does not exceed $512$ tokens, which is the first of three sub-passages used for generating question-answer pairs.
The generated pairs can be seen at the bottom of the figure.
Questions and answers highlighted in red are those that satisfy the filtering heuristics detailed in \S\ref{subsub:filtering}.
These are then translated into other languages.
\begin{center}
\noindent\fbox{%
    \parbox{0.98\textwidth}{%
    \small 
    \vspace{0.3em}
        \textbf{Title}: An American in Paris\\
        \\
        \textbf{URL}: \url{https://en.wikipedia.org/wiki?curid=309} \\
        \\
        \textbf{Text}: {\hl{An American in Paris is a jazz-influenced orchestral piece by American composer George Gershwin written in 1928. It was inspired by the time that Gershwin had spent in Paris and evokes the sights and energy of the French capital in the 1920s. Gershwin composed ``An American in Paris'' on commission from conductor Walter Damrosch. He scored the piece for the standard instruments of the symphony orchestra plus celesta, saxophones, and automobile horns. He brought back some Parisian taxi horns for the New York premiere of the composition, which took place on December 13, 1928 in Carnegie Hall, with Damrosch conducting the New York Philharmonic. He completed the orchestration on November 18, less than four weeks before the work's premiere. He collaborated on the original program notes with critic and composer Deems Taylor.} Gershwin was attracted by Maurice Ravel's unusual chords, and Gershwin went on his first trip to Paris in 1926 ready to study with Ravel. After his initial student audition with Ravel turned into a sharing of musical theories, Ravel said he could not teach him, saying, "Why be a second-rate Ravel when you can be a first-rate Gershwin?" While the studies were cut short, that 1926 trip resulted in a piece entitled ``Very Parisienne'', the initial version of ``An American in Paris'', written as a `thank you note' to Gershwin's hosts, Robert and Mabel Shirmer. Gershwin called it ``a rhapsodic ballet''; it is written freely and in a much more modern idiom than his prior works. Gershwin strongly encouraged Ravel to come to the United States for a tour. To this end, upon his return to New York, Gershwin joined the efforts of Ravel's friend Robert Schmitz, a pianist Ravel had met during the war, to urge Ravel to tour the U.S. Schmitz was the head of Pro Musica, promoting Franco-American musical relations, and was able to offer Ravel a \$10,000 fee for the tour, an enticement Gershwin knew would be important to Ravel. ...}
        \vspace{0.7em}
        \hrule
        \vspace{0.7em}
        \sethlcolor{pink} 
        \hl{\textbf{Question}: When was an American in Paris written?} \\
        \hl{\textbf{Answer}: 1928}\\
        \hl{\textbf{Type}: Number}\\ \\
        \hl{\textbf{Question}: When did George Gershwin write an American in Paris?}\\
        \hl{\textbf{Answer}: the 1920s}\\
        \hl{\textbf{Type}: Contains number}\\ \\
        \hl{\textbf{Question}: Who was the conductor of ``An American in Paris''?}\\
        \hl{\textbf{Answer}: Walter Damrosch}\\
        \hl{\textbf{Type}: Who}\\ \\
        \textbf{Question}: What was the name of the instrument that Gershwin scored for?\\
        \textbf{Answer}: automobile horns\\ \\
        \textbf{Question}: What was the name of the orchestral piece Gershwin composed in 1928?\\
        \textbf{Answer}: New York Philharmonic\\ \\
        \hl{\textbf{Question}: When did Gershwin complete the orchestration of ``An American in Paris''?}\\
        \hl{\textbf{Answer}: November 18}\\
        \hl{\textbf{Type}: Date}\\ \\
        \hl{\textbf{Question}: Who did Gershwin collaborate on the original program notes with?}\\
        \hl{\textbf{Answer}: Deems Taylor}\\
        \hl{\textbf{Type}: Who}
    }%
    \vspace{-0.3em}
}
\end{center}
\captionof{figure}{Example English question-answer pairs (on the bottom) generated from the highlighted text (in yellow) in the passage. The highlighted question-answer pairs (in red) are those that were kept after filtering.}
\label{fig:data-aug-example}


\newpage 
\section{MixCSE Loss}
\label{app:mixcse}

The MixCSE loss described in \ref{sub:mdpr_mxicse} is given by:

\begin{equation*}
   \mathcal{L}_{\text{mixcse}} = - \log \frac{\exp (\cos( \mathbf{e}_{q_i},  \mathbf{e}_{p_i^+} ) / \tau )}{\exp (\cos( \mathbf{e}_{q_i},  \mathbf{e}_{p_i^+} ) / \tau )  + \sum_{j=1}^n \exp (\cos( \mathbf{e}_{q_i},  \mathbf{e}_{p_{i,j}^-})  / \tau ) + \exp (\cos( \mathbf{e}_{q_i},  \text{SG}(\tilde{\mathbf{e}}_i))  / \tau )}
\end{equation*}

where $\tau$ is a fixed temperature and SG is the stop-gradient operator, which prevents backpropagation from flowing into the mixed negative ($\tilde{\mathbf{e}_i}$). 

